\documentclass[runningheads]{llncs}

\usepackage[T1]{fontenc}
\usepackage{cite}
\usepackage{amsmath,amsfonts}
\usepackage{algorithm}
\usepackage[noend]{algorithmic}
\usepackage{array}
\usepackage[caption=false,font=normalsize,labelfont=sf,textfont=sf]{subfig}
\usepackage{textcomp}
\usepackage{stfloats}
\usepackage{url}
\usepackage{verbatim}
\usepackage{graphicx}
\usepackage{amsmath}
\usepackage{eucal}
\usepackage{xcolor}
\usepackage{bm}
\usepackage{multirow}
\usepackage{threeparttable}
\usepackage{tabularx}
\usepackage{booktabs}
\usepackage[multiple]{footmisc}
\usepackage[colorlinks,
            linkcolor={black},
            citecolor={blue},
            urlcolor={blue},
            pdfauthor={No Author Given},
            bookmarks=false,
            pdfsubject={Random Similarity Isolation Forests},
            pdfkeywords={outlier detection, multi-modal data, distance measures, isolation forest, anomaly prediction}
            ]{hyperref}
\usepackage{xurl}

\newcolumntype{R}{>{\raggedleft\arraybackslash}X}%

\begin{document}

\title{Random Similarity Isolation Forests}

\author{Sebastian Chwilczyński\orcidID{0009-0004-1894-9131} \and
Dariusz Brzezinski\orcidID{0000-0001-9723-525X}}
\institute{Institute of Computing Science, Poznan University of Technology, Poland\\\email{dariusz.brzezinski@cs.put.poznan.pl}}

\maketitle

\begin{abstract}
With predictive models becoming prevalent, companies are expanding the types of data they gather. As a result, the collected datasets consist not only of simple numerical features but also more complex objects such as time series, images, or graphs. Such multi-modal data have the potential to improve performance in predictive tasks like outlier detection, where the goal is to identify objects deviating from the main data distribution. However, current outlier detection algorithms are dedicated to individual types of data. Consequently, working with mixed types of data requires either fusing multiple data-specific models or transforming all of the representations into a single format, both of which can hinder predictive performance. In this paper, we propose a multi-modal outlier detection algorithm called Random Similarity Isolation Forest. Our method combines the notions of isolation and similarity-based projection to handle datasets with mixtures of features of arbitrary data types. Experiments performed on 47 benchmark datasets demonstrate that Random Similarity Isolation Forest outperforms five state-of-the-art competitors. Our study shows that the use of multiple modalities can indeed improve the detection of anomalies and highlights the need for new outlier detection benchmarks tailored for multi-modal algorithms.

\end{abstract}

\section{Introduction}
\label{sec:introduction}
Outlier detection, sometimes also referred to as anomaly detection, aims to identify unusual instances that deviate from the core data distribution~\cite{aggarwal2017introduction}. Practical applications of outlier detection include preventing cyber attacks~\cite{shilay2017catching}, detecting fraud~\cite{dou2020enhancing}, and predictive maintenance~\cite{fan2022incorporating}. Over the years, specialized detection methods have been created for scalar~\cite{aggarwal2017introduction}, time series~\cite{ITURRIA2022109823}, and graph data~\cite{liu2022bond}.

However, the existing methods are usually tailored to specific types of data, whereas there is an increasing number of applications where the available data come in multiple modalities. Consider, for example, industrial equipment that has access to information from physical sensors, camera images, GPS, and communication with neighboring devices. Currently, to detect anomalies in the functioning of such equipment, one has three basic options. The first option is to focus on a single modality, risking the omission of some anomalies due to the lack of context information. The second option is to fuse the responses of multiple modality-specific detectors, which has the downside of missing anomalies that are a result of a combination of relatively normal individual sensor readings. Finally, the third option is to transform all the different modalities into a single (usually numeric) data representation. This last option is often non-trivial and has the drawback of creating very high-dimensional representations, which can hurt detection performance.

In this paper, we propose Random Similarity Isolation Forest (RSIF), an algorithm capable of detecting outliers in mixed-type datasets without having to omit any features, fuse multiple detectors, or transform the original data. Our method is a generalization of Isolation Forests~\cite{liu2008isolation} that uses similarity projections~\cite{piernik2022random} to achieve data type flexibility. We summarize our contributions as follows:
\begin{description}
    \item[New outlier detection method for multi-modal data] To our knowledge, RSIF is the first outlier detection method capable of handling mixed-type data inherently without converting it to a different representation.
    \item[Flexibility in tackling various problems] We show that our method works well regardless of whether the tackled problem comprises simple numerical data, complex objects, or mixed-type multi-modal descriptions. We validate our method on 47 datasets against five competitor models.
    \item[Evaluation of multiple similarity metrics] For different data modalities, we test a variety of similarity functions. Our study highlights the importance of selecting appropriate data projections for unsupervised anomaly detectors.
\end{description}

\section{Related work}
\label{sec:related_work}
Today, there is a plethora of outlier detection algorithms available in the literature~\cite{aggarwal2017introduction}. Among classic unsupervised outlier detection methods are proximity-based algorithms~\cite{breunig2000lof}, which are based on local neighborhood information around each data point. The most popular algorithm from this group, called Local Outlier Factor (LOF)~\cite{breunig2000lof}, calculates the distances to nearest neighbors to estimate local densities and deems data points with a substantially lower local density as outliers. Another popular group of algorithms consists of statistical models~\cite{goldstein2012histogram,li2022ecod} that first determine the probability distribution of the data and then detect outliers as points outside of the fitted distribution. Popular methods from this group include Histogram-Based Outlier Score (HBOS)~\cite{goldstein2012histogram}, which uses multiple univariate feature histograms to score outlying objects, and the recently proposed Empirical Cumulative Distribution-based Outlier Detection algorithm (ECOD)~\cite{li2022ecod}, which derives a cumulative distribution function to locate rare events in the tails of a distribution. Finally, Isolation Forest~\cite{liu2008isolation} is an example of an ensemble outlier detection method~\cite{LazarevicBagging} that detects anomalies by measuring the number of tree splits required to isolate a data point from others. However, these popular outlier detection algorithms are tailored for scalar data and cannot handle mixed-type datasets out of the box.

Datasets consisting of data points from different domains are sometimes referred to as \textit{multi-modal}. This term is used, for example, in biology when referring to multiomics patient data, in robotics when referring to data coming from different sensors, and recently in text-image-audio applications of large language models (LLMs). There have been proposals of outlier detection methods for multi-modal data, however, they either assume that the different domains all have the same representation~\cite{wellhausen2020safe} or fuse multiple data-specific models into one predictor~\cite{ide2017multi}. The proposed RSIF algorithm constructs a single model capable of handling datasets with mixtures of features of arbitrary data types.

The method presented in this paper builds upon recent advancements in classification algorithms for complex data. Sathe and Aggarwal~\cite{sathe2017similarity} have proposed a method that is capable of handling complex objects in situations where regular features are absent, but similarities between examples are attainable. The algorithm, called Similarity Forest, uses a distance-based projection to order objects and create decision tree splits. More recently, Piernik et al. have proposed the Random Similarity Forest algorithm~\cite{piernik2022random}, a combination of Similarity Forest and Random Forest~\cite{breiman2001random} capable of handling scalar, complex, and mixed datasets. However, the aforementioned algorithms are supervised learners and require labeled data to perform similarity-based projections. In this paper, we combine ideas from projection-based algorithms with the notion of isolation known from Isolation Forests~\cite{liu2008isolation}. Our approach proposes a new way of selecting reference objects for projections to create an unsupervised outlier detection algorithm suitable for multi-modal problems.

\section{Random Similarity Isolation Forest}
\label{sec:algorithm}
\begin{figure*}[htb]
\centering
\includegraphics[width=0.95\textwidth]{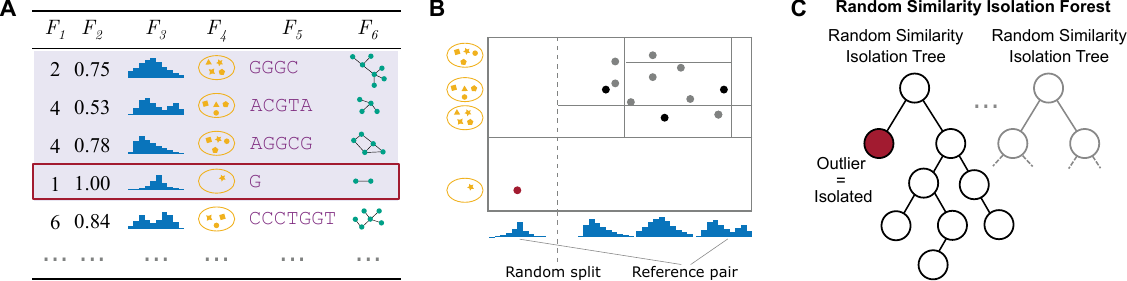}
\caption{Example of training a Random Similarity Isolation Forest. (\textbf{\textsf{A}}) The input dataset with six features ($F_1$--$F_6$) of different types (categorical, numeric, histogram, set, sequence, graph). The dark background rows symbolize the subsample going into an example tree. (\textbf{\textsf{B}}) Using reference pairs, distance-based projections are used to map data points. In the projection space, random splits are performed to create trees and isolate outliers.(\textbf{\textsf{C}}) A schematic depiction of a Random Similarity Isolation Tree.}
\label{fig:rsif_example}
\end{figure*}

In the following sections, we introduce the used notation, explain the proposed Random Similarity Isolation Forest (RSIF) algorithm, and discuss how reference objects affect similarity projections. A schematic overview of the Random Similarity Isolation Forest algorithm is presented in Figure~\ref{fig:rsif_example}.

\vspace{-1cm}

\subsection{Notation}
\label{sec:algorithm:notation}
By $\mathcal{X} = \{\bm{x_1}, \bm{x_2},..., \bm{x_n} \}$ we will denote a dataset of $n$ examples, where every example $\bm{x_i}=(x_{i_1}, x_{i_2},... x_{i_p})$ consists of $p$ feature values. The dataset's $k$-th feature vector (`column') will be denoted by $F_k=(x_{1_k}, x_{2_k},... x_{n_k})$, with $F_k \in \mathcal{F}$, where $\mathcal{F}$ is the set of all features. We note that the features do not have to be numerical but can consist of arbitrary objects (Fig.~\ref{fig:rsif_example}A). Moreover, $\Delta_k = \{ \delta: F_k \times F_k\ \mapsto \mathbb{R} \}$ will denote a set of distance measures defined for feature $F_k$.

Given $\mathcal{X}$, the task of unsupervised outlier detection is to derive a function \emph{f}: $\mathcal{X} \mapsto \mathbb{R}$ returning an outlier score for every, possibly previously unseen, example $\bm{x_i}$. These scores provide a ranking
that reflects the degree of anomaly. A threshold  $\theta$ can be selected so  that if $f(\bm{x_i}) \ge \theta$ then $\bm{x_i}$ is classified as an outlier.

\subsection{The algorithm}
\label{sec:algorithm:pseudocode}
Analogously to Isolation Forest~\cite{liu2008isolation}, the Random Similarity Isolation Forest (RSIF) algorithm builds a forest of isolation trees. There are three hyperparameters to the algorithm: the number of trees~$t$, subsampling size $\psi$, and a vector of sets of distance measures $\bm{\delta}= (\Delta_1, \Delta_2,..., \Delta_p)$ (each feature can have many applicable distance measures). Parameter $t$ determines the size of the forest, $\psi$ controls the per-tree training data size, and $\bm{\delta}$ defines the distance measures that can be used for projecting each of the features in the dataset. The pseudocode for the forest building procedure is presented in Algorithm~\ref{alg:forest}.
\vspace{-0.5cm}

\begin{algorithm}[htb]
    \caption{RSIF: Random Similarity Isolation Forest}
    \label{alg:forest}
\begin{algorithmic}[1]
    \REQUIRE $\mathcal{X}$: input data; $t$: number of trees; $\psi$: subsampling size; $\bm{\delta}$:~vector of sets of distance measures for each feature.\\
    \ENSURE A set of Random Similarity Isolation Trees.
    \STATE $\mathit{forest} \leftarrow \emptyset$
    \STATE $\mathit{max\_depth} \leftarrow \lceil\log_2\psi\rceil$
    \FOR{$i \leftarrow 1, 2, \ldots, t$}
        \STATE $\mathcal{X}_i \leftarrow \mathit{subsample}(\mathcal{X}, \psi)$
        \STATE $\mathit{forest} \leftarrow \mathit{forest} \cup \mathit{RSIT}(\mathcal{X}_i, \bm{\delta}, \mathit{max\_depth})$
    \ENDFOR
    \RETURN{$\mathit{forest}$}
\end{algorithmic}
\end{algorithm}
\vspace{-0.5cm}

A forest consists of $t$ Random Similarity Isolation Trees (Algorithm~\ref{alg:tree}). To build a tree, the algorithm uses a subset of the available data $\mathcal{X}_s\subseteq\mathcal{X}$. If for $\mathcal{X}_s$ the stopping conditions are met (max tree depth reached or no pair of objects with a distance greater than zero for any feature), the tree is finished (lines~\ref{l:stop}--\ref{l:stop:end}). Otherwise, the algorithm selects a random feature $F_k$, a distance measure $\delta$ suitable for the selected feature, and a pair of reference objects $\bm{x_q},\bm{x_r}\in\mathcal{X}_s$ (lines~\ref{l:f_pick}--\ref{l:p_pick:end}). With all the above ingredients, the algorithm performs a distance-based projection to create a dynamic feature vector $P$ (lines \ref{l:proj}--\ref{l:proj:end}) and makes a random split at $\mathit{thr}$, i.e., at a point within the range of the calculated dynamic feature vector (line~\ref{l:thr}). The split produces two subsets of the data ($\mathcal{X}_{\mathit{left}}$, $\mathcal{X}_{\mathit{right}}$), which are recursively used to create subsequent splits (lines~\ref{l:split:fl}--\ref{l:node:r}).

\begin{algorithm}[htb]
    \caption{RSIT: Random Similarity Isolation Tree}
    \label{alg:tree}
\begin{algorithmic}[1]
    \REQUIRE $\mathcal{X}_s$: (sub)tree training data; $\bm{\delta}$: vector of sets of distance measures for each feature; \textit{max\_depth}: maximal tree depth.\\
    \ENSURE A trained Random Similarity Isolation Tree model.
    \IF{$\mathit{depth \geq max\_depth}$ \OR $\neg\exists_{k, \delta, x_{i_k}, x_{j_k}}\delta(x_{i_k}, x_{j_k}) > 0 $}\label{l:stop}
        \STATE{$\mathit{setLeaf()}$}
        \RETURN{$\mathit{self}$}\label{l:stop:end}
    \ELSE
        \STATE $\mathcal{F}_{\delta > 0} \leftarrow \{F_k\in\mathcal{F}\mid \exists {\delta \in \bm{\delta}[k], x_{i_k}, x_{j_k}}:\delta(x_{i_k}, x_{j_k}) > 0\}$
        \STATE{$F_k\leftarrow \mathit{random}(\{F_k\in\mathcal{F}_{\delta > 0}\})$}\label{l:f_pick} {\hfill$\triangleright$ Random feature $k$}
        \STATE $\delta \leftarrow \mathit{random}(\{\delta : \delta \in \bm{\delta}[k]\})$ {\hfill$\triangleright$ Random distance}
        \STATE{$x_{u_k} \leftarrow \mathit{random}(\mathcal{X}_s)$ } \label{l:p_pick:start}
        \STATE{$x_{q_k} \leftarrow \underset{x_{i_k}}{\text{argmax}}\,\delta(x_{i_k}, x_{u_k})$}{\hfill$\triangleright$ Reference objects $q$, $r$} 
        \STATE{$x_{r_k} \leftarrow \underset{x_{i_k}}{\text{argmax}}\,\delta(x_{i_k}, x_{q_k})$} \label{l:p_pick:end}
        \FOR{$\bm{x_i} \in \mathcal{X}_s$} \label{l:proj}
            \STATE{$P(x_{i_k})\leftarrow\delta(x_{r_k}, x_{i_k}) - \delta(x_{q_k}, x_{i_k})$}
        \ENDFOR\label{l:proj:end}
        \STATE$\mathit{thr}\leftarrow  \mathit{random}(\{ x: x \in [\mathit{min(P)};\mathit{max(P)}] \})$\label{l:thr}
        \STATE{$\mathcal{X}_{\mathit{left}}\leftarrow \{\bm{x_i}\in\mathcal{X}_s: P(x_{i_k}) \leq \mathit{thr}\} $}\label{l:split:fl}
        \STATE{$\mathcal{X}_{\mathit{right}}\leftarrow \{\bm{x_i}\in\mathcal{X}_s: P(x_{i_k}) > \mathit{thr}\} $}\label{l:split:fr}
        \STATE{$N_{\mathit{left}}\leftarrow \mathit{RSIT}(\mathcal{X}_{\mathit{left}}, \bm{\delta}, \mathit{max\_depth})$}\label{l:node:l}
        \STATE{$N_{\mathit{right}}\leftarrow \mathit{RSIT}(\mathcal{X}_{\mathit{right}}, \bm{\delta}, \mathit{max\_depth})$}\label{l:node:r}
    \ENDIF

    \RETURN{$\mathit{self}$}
\end{algorithmic}
\end{algorithm}

The essential component making our approach multi-modal is the use of distance-based projections as dynamic features. We assume that for all $F_k$ there exists a multidimensional space $\mathcal{V} \subseteq \mathbb{R}^n$ in which the objects can be embedded. Sathe and Aggarwal~\cite{sathe2017similarity} prove that a projection $P(x_{i_k})$ of $\bm{x_i}$ into a direction defined by $x_{q_k}$ and $x_{r_k}$ can be approximated with:
\begin{equation}
P(x_{i_k}) = \delta(x_{r_k}, x_{i_k}) - \delta(x_{q_k}, x_{i_k})
\label{eq:projection}
\end{equation}
Since the projection is different for each selected pair of reference objects, it can be regarded as a dynamic feature. Moreover, nothing restricts one from defining many distance functions for one $F_k$. Then, each distance function selected for $F_k$ will create a separate feature. Each of them may highlight different characteristics of complex objects and increase the algorithm's discriminative power.

After all isolation trees are created, for an example $\bm{x}$ an anomaly score $\emph{f}(\bm{x})$ is calculated in the same way as in Isolation Forests~\cite{liu2008isolation}:
$\emph{f}(\bm{x}) = 2^{(-\frac{E(h(\bm{x}))}{c})}$, where $E(h(\bm{x}))$ is the length of a path from a root node to a leaf containing $\bm{x}$ averaged over all trees in a forest, and $c$ is the average path length of the trees.

The computational complexity of Random Similarity Isolation Forest is $\mathcal{O}(d \cdot t \cdot \psi \cdot \log \psi)$, where $d$ is the cost of distance calculation, $t$ is the number of trees, and $\psi$ is the sub-sampling size. This can be optimized to $\mathcal{O}(\frac{mnd}{2} + t \cdot \psi \cdot \log \psi)$ by restricting potential reference objects to a subset of $m < n$ objects and precomputing an $m \times n$ distance matrix. The space complexity is $\mathcal{O}(t \cdot 2^{\lceil\log_2\psi\rceil} + |\Delta| \cdot m \cdot n)$, where $|\Delta|$ is the number of distance functions.

\subsection{Reference Objects Selection}
\label{sec:algorithm:reference}

One of the key elements of the proposed algorithm is the selection of reference objects that will be used to project examples onto a dynamic feature (Algorithm~\ref{alg:tree}, lines~\ref{l:p_pick:start}--\ref{l:p_pick:end}). To highlight the importance of this step, let us illustrate how a distance-based projection $P$ (Eq.~\ref{eq:projection}) works depending on reference objects.

\begin{figure}[htb]
    \centering
    \includegraphics[width=0.8\linewidth]{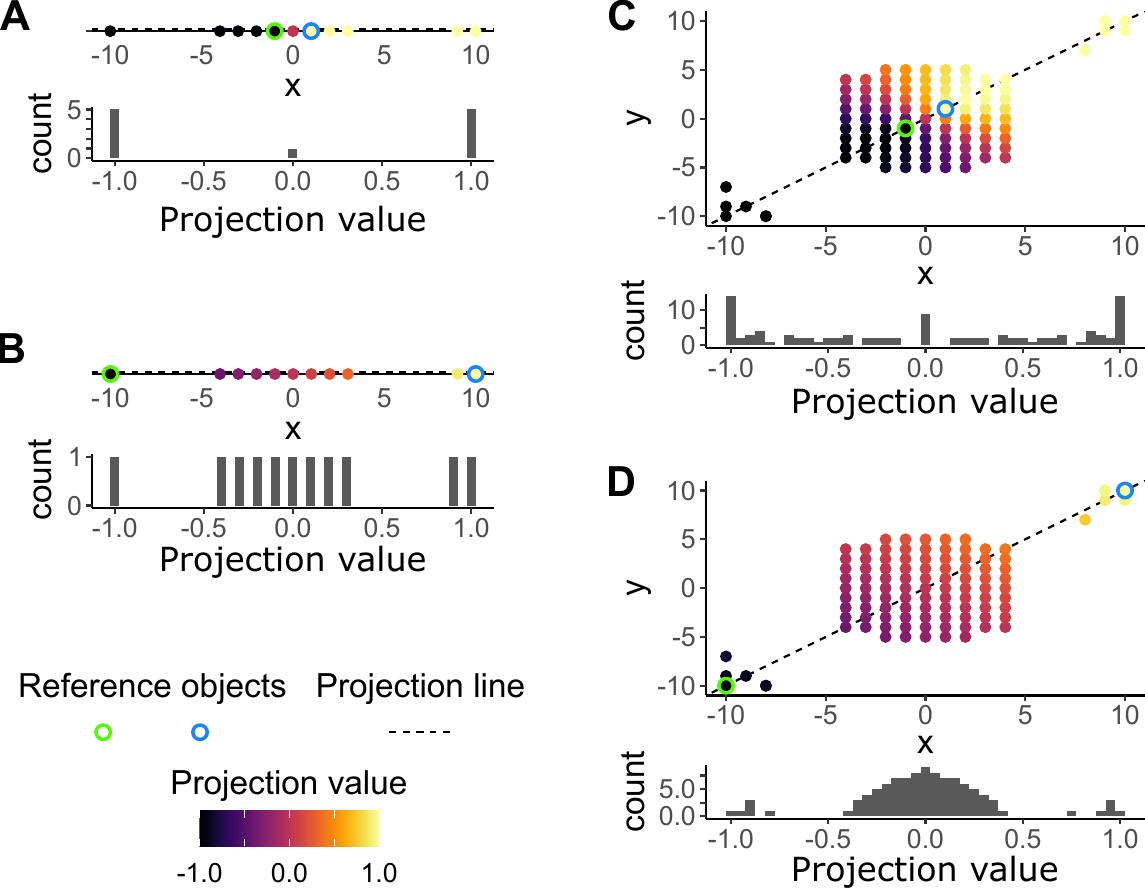}
    \caption{Example distance based projections in 1D (A, B) and 2D (C,D) Euclidean space. The top part of each subfigure presents examples (objects) represented as points. Each point is colored according to the value obtained after its projection. The bottom figure presents the distribution of projection values.}
    \label{fig:reference_examples}
\end{figure}

Figure~\ref{fig:reference_examples} presents four example projections of objects described by a feature $F_k$ consisting of one (subfigures A, B) or two numeric values (C, D). The distance $\delta$ used for the projection is the Euclidean distance. The projection values were calculated according to Eq.~\ref{eq:projection} and then 0--1 normalized to show all the examples on the same color scale.

Let us first notice that if $\delta(x_{q_k}, x_{r_k}) = 0$, then, for any $x_{i_k}$, $\delta(x_{r_k}, x_{i_k}) = \delta(x_{q_k}, x_{i_k})$ and projection $P(x_{i_k}) = 0$. Therefore, projections are meaningless if there is no distance between the reference objects because all examples would get the same projection value. Comparing~Fig.~\ref{fig:reference_examples}A against Fig.~\ref{fig:reference_examples}B, it can also be noticed that the concrete distance between the reference objects is also important. When the reference objects are close to each other, many examples become indiscernible because they get the same projection value. In the 1D case, any point `behind' a reference object gets the same projection value as the closest reference (Fig.~\ref{fig:reference_examples}A). Since outliers are often expected to reside at the extremes of the data distribution~\cite{goldstein2012histogram,li2022ecod}, reference objects should be far apart to make outlying values separable from the rest of the examples (Fig.~\ref{fig:reference_examples}B).

A similar situation can be noticed in the more complicated case of 2D Euclidean distance. If the reference objects are very close to each other (Fig.~\ref{fig:reference_examples}C), then many examples at the extremes of the distribution are indiscernible (see the histogram peeks at -1 and 1 in the bottom panel of Fig.~\ref{fig:reference_examples}C). On the other hand, when the reference objects are at the extremes (Fig.~\ref{fig:reference_examples}D), all the points are discernible, and the distribution of projection values is most likely much more suitable for outlier detection. However, for more complicated spaces and distances, there might not be one global or local best choice of reference objects. That is why we propose a two-step approach that selects a random object $x_{i_k}$, then the furthest from it $x_{q_k}$ and the furthest from that $x_{r_k}$ (Algorithm~\ref{alg:tree}, lines~\ref{l:p_pick:start}--\ref{l:p_pick:end}). This way, we aim to select the two furthest objects on one of the most promising projection lines. This two-step approach will be compared against selecting a local optimum, global optimum, and a random pair in Section~\ref{sec:experiments:sensitivity}.

We note that the original projections proposed in Similarity Forests~\cite{sathe2017similarity,piernik2022random} used reference objects selected from two different classes. The goal therein was to create a split that separates examples from opposing classes. Therefore, the fact that examples within a class are indiscernible after the projection was not a problem as long as the split between the two classes was correct. In this paper, we tackle the problem of unsupervised outlier detection and there are no class labels. That is why the reference object selection proposed in previous papers~\cite{sathe2017similarity,piernik2022random} does not apply to our setting and why the proposed procedure constitutes a major contribution of this paper.

\section{Experiments}
\label{sec:experiments}

\subsection{Experimental Setup}
\label{sec:experiments:setup}
We compare RSIF against five outlier detection methods: Isolation Forest (IF)~\cite{liu2008isolation}, Local Outlier Factor (LOF)~\cite{breunig2000lof}, Histogram Outlier Score (HBOS)~\cite{goldstein2012histogram}, Empirical-Cumulative-distribution-based Outlier Detection (ECOD)~\cite{li2022ecod}, and Similarity Forest (SF)~\cite{sathe2017similarity}. The first four methods (IF, LOF, HBOS, ECOD) are among the most popular general-purpose outlier detectors. Among these methods, ECOD is the most recent approach and has been shown to perform favorably in benchmarks~\cite{li2022ecod}. Similarity Forest was added to show the difference between treating examples as complex objects rather than mixtures of features.

The experiments aimed to comprehensively test various distance measures that could be used with LOF, SF, and RSIF. We used traditional measures for scalars and numeric vectors, such as Euclidean, manhattan, Chebyshev, and cosine distance. For categorical data, we tested~\cite{boriah2008similarity}: Goodall, Lin, and occurrence frequency. For graphs, we additionally used adjacency matrix measures~\cite{hartle2020network}: portrait divergence, degree divergence, and NetLSD. Moreover, we tested Wasserstein distance~\cite{villani2009optimal} for histograms and dynamic time warping (DTW)~\cite{berndt1994using} for time series. Finally, a special distance measure denoted as \textit{identity} signified no projection, i.e., using the same order of examples as would be used by IF.

Unlike RSIF and SF, all other compared algorithms require numerical representation of the data. While most data types can be easily represented numerically without requiring additional processing, the handling of graphs and sequences of sets demands special care. To address this, we adopted the bag-of-words (BoW) representation. For graphs, a binary vector $\bm{v_i}$ of length $|\mathcal{V}|$ is created, where $\mathcal{V} = \bigcup{V_i}$, and $V_i$ represents the set of vertices for the $i$-th graph in a particular dataset. Sequences of sets were treated analogously, but instead of considering a set of all nodes, we examined a set of all possible elements.

The experiments were implemented in Python 3.9. For experiments involving IF, LOF, HBOS, and ECOD, we used their implementations from the scikit-learn~\cite{scikit-learn} and PyOD~\cite{zhao2019pyod} libraries. The SF algorithm was modified to use RSIF's two-step reference pair selection method proposed in this paper. Unless stated otherwise, we relied on the default hyperparameters for each algorithm, as specified in their implementations. In each experiment, a stratified sample of 70\% of the data was used for training, with the remaining 30\% set aside for testing. Performance was evaluated by taking the average score of 10 independent trials using average precision (AP) and the area under the ROC curve (AUC). To compare the obtained results, we performed the non-parametric Friedman statistic test with a post-hoc Connover test to distinguish whether some algorithms perform statistically better than others.

\subsection{Datasets}
\label{sec:experiments:datasets}
We used 47 datasets taken from outlier detection benchmarks and generators for different types of data.%
\footnote{\scriptsize\url{https://github.com/Minqi824/ADBench}}
\footnote{\scriptsize\url{https://outlier-detection.github.io/utsd}}
\footnote{\scriptsize\url{https://gingerbread.shinyapps.io/SequencesOfSetsGenerator/}}
\footnote{\scriptsize\url{https://github.com/GuansongPang/ADRepository-Anomaly-detection-datasets}} All the datasets can be found in the code repository accompanying this paper: \url{https://github.com/SebChw/RSIF}. When looking for benchmarks, we favored those in which the examples marked as outliers constituted 5\% or fewer examples in the dataset. The datasets were divided into four groups: \textit{sensitivity analysis} (independent datasets for analyzing the properties of RSIF), \textit{scalar data}, \textit{complex data}, \textit{mixed data} (multimodal data). A detailed description of the datasets can be found in the repository.\footnote{\scriptsize Detailed dataset descriptions, sensitivity test results, supplementary plots, and the AUC ROC results are available at the repository: \url{https://github.com/SebChw/RSIF}}

\subsection{Sensitivity analysis}
\label{sec:experiments:sensitivity}
During our analyses of Random Similarity Isolation Forests (RSIF), we investigated the impact of: the number of trees $t \in \{1,5,10,25,50,100,200\}$,
the subsample size $\psi \in \{64,128,256,512\}$,
the ratio of examples available as reference objects $m \in \{0.10, 0.25, 0.50, 0.75, 1.00\}$,
the effect of the reference pair selection strategy,
the used distance measures $\bm{\delta}$.

The analysis of parameters $t$ and $\psi$ revealed that RSIF behaves similarly to IF. Increasing the number of trees in the ensemble improves its performance up to a certain saturation point. Hence, we chose $t=100$ as the default, just as is the case for IF~\cite{liu2008isolation}. The effect of the subsample size $\psi$ was also consistent with IF. Therefore, we use the same default $\psi=256$~\cite{liu2008isolation}. Finally, the way the size of the reference objects pool $m$ affected RSIF's performance varied slightly between datasets (Fig.~\ref{fig:hyperparameters}). That is why, in our experiments, we use $m=0.5$, as a safe option that still computes distances twice as fast.

As mentioned in Section~\ref{sec:algorithm:reference}, the way in which reference objects are selected can play a role in how RSIF works. Therefore, we evaluated the effectiveness of four reference pair selection strategies: \textit{random}, \textit{global}, \textit{local}, and \textit{two-step}. The \textit{random} strategy selected reference objects randomly, \textit{global} selected 10 most distant pairs from the entire training dataset and then sampled them randomly, \textit{local} selected the most distant pair from an individual tree's subsample, whereas \textit{two-step} selected reference objects as presented in Algorithm~\ref{alg:tree} (lines~\ref{l:p_pick:start}--\ref{l:p_pick:end}). The differences between the strategies varied between datasets and used distance measures, with the \textit{two-step} strategy achieving the best average rank: $r_{\mathit{two\-step}}=2.35$, $r_{\mathit{global}}=2.43$, $r_{\mathit{random}}=2.53$, $r_{\mathit{local}}=2.68$ (lower is better).

We have also studied how the selection of the number and type of distance measures used by RSIF affects its performance (Fig.~\ref{fig:distances}). Observing the numeric datasets, we see that the best distance is usually not only data type-dependent but also problem-dependent. For example, degree divergence plays a key role in detecting outliers in the \texttt{bzr} dataset but was not in the top-3 set of measures for \texttt{cox2} or \texttt{dhfr}. It is clear that there is no best default set of distance measures, and the set of used distances should be selected by an expert on a use-case basis. Not having the expertise to select distance measures for every test dataset (and to avoid the risk of cherry-picking measures to obtain the best final results), for the experimental comparison with competitive methods, all the distance-based algorithms (LOF, SF, RSIF) had their distance measures selected based on a 30\% validation portion within the training set.

\subsection{Comparison with existing methods}
\label{sec:experiments:results}
Table~\ref{tab:datasets} presents the Average Precision (AP) results of the experimental comparison of Random Similarity Isolation Forest (RSIF) against the five 
analyzed competitive methods: Isolation Forest (IF), Local Outlier Factor (LOF), Histogram-based Outlier Score (HBOS), Empirical-Cumulative-distribution-based Outlier Detection (ECOD), and Similarity Forest (SF).

As can be noticed by looking at Table~\ref{tab:datasets}, the winning algorithms are different for different types of data. RSIF is usually the best when it comes to numerical, graph, and multiomics data, ECOD and HBOS perform well for categorical and sequential data, whereas LOF is particularly good for time series and (text and image) vector embeddings. 
The AP differences between algorithms are statistically significant according to the Friedman test, with RSIF achieving the best average rank (2.81) and being statistically significantly better than IF, ECOD, and HBOS according to the Connover post-hoc test. Similar rankings were observed for AUC (Table~\ref{tab:datasets_AUC}).

\begin{table}[H]
\scriptsize
\centering
\caption{Average Precision (AP) performance of RSIF and five competitive methods. The best results on each dataset are highlighted in bold, and the second best are underlined. Avg. rank presents the Friedman test rank (lower is better).}
\begin{tabularx}{\textwidth}{l@{}ccrrcXXXXXX}
\toprule
\multirow{2.5}{*}{Dataset} & \multirow{2.5}{*}{Type} & \multirow{2.5}{*}{Features} & \multirow{2.5}{*}{\#Ex.} & \multirow{2.5}{*}{\#Feat.} & \multirow{2.5}{*}{\%Outlier} & \multicolumn{6}{c}{AP}\\
\cmidrule{7-12}
& & & & & & IF & LOF & HBOS & ECOD & {\,\,SF} & RSIF \\
\midrule
\texttt{glass} & \multirow{16}{*}{scalar}  & \multirow{11}{*}{numeric} & 214 & 7 & 4.21 &0.13&0.15&\underline{0.15}&0.12&0.14&\textbf{0.17}\\
\texttt{letter} & & & 1600 & 32 & 6.25 &0.09&\textbf{0.57}&0.09&0.08&0.22&\underline{0.24}\\
\texttt{musk} & & & 3062 & 166 & 3.17 &  0.96&0.22&1.00&0.51&\textbf{1.00}&\underline{1.00} \\
\texttt{annthyroid} & & & 7200 & 6 & 7.42 & \underline{0.27}&0.17&0.22	&0.26&0.25&\textbf{0.36}\\
\texttt{satimage} & & & 5803 & 36 & 1.22 &0.93&0.01&0.77&0.67&	\underline{0.94}&\textbf{0.94} \\
\texttt{thyroid} & & & 3772 & 6 & 2.47 & 0.55&0.02&0.53&0.52&\underline{0.58}&\textbf{0.64}\\
\texttt{vowels} & & & 1456 & 12 & 3.43 &  0.09&\textbf{0.35}&0.09&0.09&0.06&\underline{0.33} \\
\texttt{waveform} & & & 3443 & 21 & 2.90 & 0.07&\underline{0.11}&0.05&0.04&	\textbf{0.45}&0.10\\
\texttt{wbc} & & & 223 & 9 & 4.48 & \textbf{0.96}&0.31&0.89&\underline{0.93}&0.84&\textbf{0.96} \\
\texttt{wdbc} & & & 367 & 30 & 2.72 & \underline{0.82}&0.65&0.80&0.49&\textbf{0.92}&0.71 \\
\texttt{wilt} & & & 4819 & 5 & 5.33 & 0.05&\textbf{0.08}&0.04&0.04&0.04&\underline{0.06}\\
\cmidrule{3-12}
\texttt{aid} & & \multirow{5}{*}{categorical} & 4278 & 114 & 1.40 &  \textbf{0.03}&0.02&0.03&0.03&0.03&\underline{0.03} \\
\texttt{apascal} & & & 12694 & 64& 1.39 &  0.01&0.02&\textbf{0.02}&\underline{0.02}&0.01&\underline{0.02} \\
\texttt{cmc} & & & 1472 & 8 & 1.97& 0.06&\textbf{0.08}&\underline{0.08}&\underline{0.08}&0.06&0.07 \\
\texttt{reuters} & & & 12896 & 100 & 1.84&  0.50&0.25&\underline{0.58}&\textbf{0.61}	&0.57&0.44 \\
\texttt{solarflare} & & & 1065 & 11 & 4.04 &  0.16&0.05&0.19&0.18&\textbf{0.25}&\underline{0.23} \\
\midrule
\texttt{nci1} & \multirow{16.2}{*}{complex} & \multirow{5}{*}{graph} & 2160 & 1 & 4.77&  0.05&\underline{0.06}&0.05&0.05&0.05&\textbf{0.06} \\
\texttt{aids} & & & 1680 & 1 & 4.76 &  0.52&0.14&0.67&0.43&\underline{0.96}&\textbf{0.97} \\
\texttt{enzymes} & & & 105 & 1 & 4.76 &  \textbf{0.32}&0.25&0.22&\underline{0.28}&0.17&0.24 \\
\texttt{proteins} & & & 696 & 1 & 4.47 &  0.06&0.11&0.05&0.09&\textbf{0.12}&\underline{0.12} \\
\texttt{dd} & & & 726 & 1 &4.82&  0.10&0.10&0.04&0.21&\underline{0.26}&\textbf{0.28} \\
\cmidrule{3-12}
\texttt{earthquakes} & & \multirow{4}{*}{time series} & 387 & 512 & 4.91 & 0.11&\underline{0.12}&0.07&0.10&0.08&\textbf{0.12} \\
\texttt{aibo} & & & 367 & 70 & 4.90 &  \textbf{0.12}&\underline{0.11}&0.10&0.10&0.06	&0.07 \\
\texttt{ECGFiveDays} & & & 465 & 136 & 4.95 & \underline{0.18}&\textbf{0.34}&0.13&0.10&0.12&0.16\\
\texttt{MPOC} & & & 583 & 80 & 4.97 & \underline{0.20}&\textbf{0.34}&0.17&0.16&0.16&0.18 \\

\cmidrule{3-12}
\texttt{amazon} & & \multirow{3}{*}{text} & 10000 & 768 & 5.00 & 0.05	& \textbf{0.06}&	0.05&	0.05&	\underline{0.05}&	\underline{0.05} \\
\texttt{imdb} & & & 10000 & 768 & 5.00&  0.05 &	\textbf{0.06} &	0.05	&0.05	&\underline{0.05}	&\underline{0.05} \\
\texttt{yelp} & & & 10000 & 768 &5.00 &0.05	& \textbf{0.07}	&0.05&	0.05	&\underline{0.05}	&\underline{0.05} \\
\cmidrule{3-12}
\texttt{cifar} & & \multirow{3}{*}{image} & 5263 & 512 & 5.00&  \underline{0.15}	& \textbf{0.16}	& 0.12	&0.13	&0.12	&0.14 \\
\texttt{fashionmnist} & & & 6315 & 512 &5.00 & \textbf{0.21}	& 0.19&	0.14&	0.18&	0.19&	\underline{0.20} \\
\texttt{svhn} & & & 5208 & 512 & 5.00 &  0.06 &	\textbf{0.10} &	0.05	&0.06	&\underline{0.08}	&0.06 \\
\midrule
\texttt{items} & \multirow{6.4}{*}{mixed} & \multirow{2}{*}{sequences} & 210 & 1 & 4.76 & \underline{0.27} &	\textbf{0.28}	&\underline{0.27}&	\underline{0.27}&	0.20	&0.23 \\
\texttt{length} & & \multirow{2}{*}{of sets} & 210 & 1 & 4.76 &  0.54 &	0.54	& \textbf{0.60}	& \underline{0.59}	&0.52	&0.45 \\
\texttt{order} & & & 210 & 1 & 4.76 &  0.09	& \underline{0.10}	&0.09&	\textbf{0.11}	&0.07	&0.09 \\
\cmidrule{3-12}
\texttt{ovarian} & & \multirow{3}{*}{multiomics} & 125 & 50 & 4.80 & 0.08	&0.06	&0.09	&\underline{0.10}	&0.06&	\textbf{0.18} \\
\texttt{breast} & & & 770 & 50 & 3.64 &  0.06	&\underline{0.20}	&0.04&	0.07&	0.07&	\textbf{0.27} \\
\texttt{rosmap} & & & 177 & 600 & 4.52 &  \textbf{0.23}	&0.14&	0.20&	\underline{0.21}&	0.18&	0.16\\
\midrule
&&&&& \textbf{Avg. rank} & 3.21 & 3.21 & 3.89 & 3.81 & 4.05 & \textbf{2.81} \\
\bottomrule
\end{tabularx}
\label{tab:datasets}
\end{table}

The above analysis shows that different algorithms are suited for different types of outliers. ECOD and HBOS detect outliers in tails of feature distributions~\cite{li2022ecod}, IF detects isolated points based on random features, LOF and SF look at global distances between objects, and RSIF, depending on the used distance, isolates points either based on individual features or global distances. To quantify how the studied algorithms related to each other, we have computed the intraclass correlation coefficient (ICC(C,1)) as a measure of inter-algorithm agreement. We found that ECOD, HBOS, and IF are very strongly correlated, LOF and SF are the most uncorrelated approaches, whereas RSIF shares similarities with all the algorithms but LOF (Fig.~\ref{fig:icc}).

Seeing that RSIF obtained the best average rank in the experiments and shares common features with different methods, it can be argued that RSIF serves as a compromise between detecting local and global outliers. However, this issue should be further studied on a wider collection of datasets and distances, especially with mixed-type features which currently have very few benchmarks.

\section{Discussion and conclusions}
\label{sec:discussion}

In this paper, we introduced Random Similarity Isolation Forest, a novel algorithm that detects outliers in mixed-type datasets using distance-based projections and random splits. Our method outperformed five existing outlier detection algorithms and uniquely handles both traditional numerical features, complex data types like graphs and time series, and mixtures of different modalities. 

The proposed Random Similarity Isolation Forest (RSIF) algorithm can be considered a generalization of Isolation Forests (IF)~\cite{liu2008isolation}. Indeed, if RSIF were to be used only for numerical data and the Euclidean distance, it would work just as IF. As the experimental results show (Table~\ref{tab:datasets}), in most cases the ability to use multiple distance measures lets RSIF achieve better performance than IF.

The biggest benefit of RSIF is its ease of use for any type of data. Just by selecting pre-existing distance measures, we were able to quickly get an outlier detector for numerical data, graphs, and multiomics. What is important, this property can be easily transferred to other algorithms. As was shown in Section~\ref{sec:algorithm:reference} (Fig.~\ref{fig:reference_examples}), distance projections create new dynamic features with their own value distributions. Such projected features can be easily applied to statistical models such as HBOS~\cite{goldstein2012histogram} or ECOD~\cite{li2022ecod}, which detect outliers based on feature histograms. The generalization of HBOS, ECOD, and other methods to complex and mixed-type data constitutes a promising line of future research. 

It is also worth noting that the presented experimental study was the first to analyze the use of Similarity Forests (SF)~\cite{sathe2017similarity}. The relatively poor performance of SF (worst average rank) provides a couple of interesting insights. First, it seems that calculating distance measures for multiple features (RSIF) rather than entire objects (SF) offers more flexibility in detecting outliers. Secondly, most outliers in the benchmark datasets are local rather than global. In other words, most of the time, a single feature or a single value in a vector makes an object an outlier rather than an unlikely combination of all the features.

To the best of our knowledge, this is the first study to combine outlier detection datasets from standard numeric benchmarks, graph repositories, time series data, and multiomics. In doing so, we found that time series outlier detection datasets are often subsampled time series classification datasets. Text, image, and multiomics data are also modified classification datasets. Graph repositories, on the other hand, focus on detecting outlying node types or node counts rather than atypical connections and network structures. To improve future studies on outlier detection for different types of data, it may be worthwhile to assemble datasets in a unified repository, clearly state the nature of outliers, and aim to gather more multi-modal datasets, which are currently underrepresented.

Finally, as future work, it would be worthwhile to investigate the pros and cons of using multi-modal representations of data in a set of case studies, for example, involving predictive maintenance tasks. With high volumes of complex and unstructured data being gathered in the industry, we believe that multi-modal anomaly detection constitutes an important research topic for the upcoming years. 

\section*{Acknowledgments}
We would like to thank Florian Lemmerich, Maciej Piernik, Wojciech Cieśla, Karol Cyganik, and Oskar Szudzik for their suggestions and input on early versions of this study. This research was funded by the Poznan University of Technology, Institute of Computing Science Statutory Funds.

\bibliographystyle{splncs04}
\bibliography{references}

\appendix
\renewcommand{\theequation}{S\arabic{equation}}
\renewcommand{\thefigure}{S\arabic{figure}}
\renewcommand{\thetable}{S\arabic{table}}
\renewcommand{\thesection}{S\arabic{section}}
\setcounter{figure}{0}    
\setcounter{table}{0}    

\section{Datasets}

The used datasets were taken from outlier detection benchmarks and generators for different types of data:
\begin{itemize}
    \item numerical and categorical data: \url{https://github.com/Minqi824/ADBench}
    \item time series: \url{https://outlier-detection.github.io/utsd}
    \item sequences of sets: \url{https://gingerbread.shinyapps.io/SequencesOfSetsGenerator/}
    \item images, text, and graphs: \url{https://github.com/GuansongPang/ADRepository-Anomaly-detection-datasets}
\end{itemize}
The exact versions of the datasets can be found in the code repository accompanying this paper. When looking for benchmarks, we favored those in which the examples marked as outliers constituted 5\% or fewer examples in the dataset.

\textbf{Sensitivity analysis.} For the analysis of the hyperparameters of RSIF, we used datasets that were independent of those used for the experimental comparison with other methods. More precisely, we used 10 datasets: 4 numerical (\texttt{cardio}, \texttt{lymphography}, \texttt{optdigits}, \texttt{speech}), 1 categorical (\texttt{ad}), 3 graph (\texttt{cox2}, \texttt{bzr}, \texttt{dhfr}), 1 text embedding (\texttt{agnews}), and 1 time-series (\texttt{twoleadecg}).

\textbf{Scalar data.} For tests with scalar data, we used 10 popular benchmark datasets, including 5 numerical (\texttt{glass}, \texttt{musk}, \texttt{satimage}, \texttt{vowels}, \texttt{wbc}) and 5 categorical (\texttt{aid}, \texttt{apascal}, \texttt{cmc}, \texttt{reuters}, \texttt{solar\-flare}). The datasets were chosen for their variety in terms of the number of examples and features (Table~\ref{tab:datasets}).

\textbf{Complex data.} For experiments with complex objects, we used 5 graph (\texttt{aids}, \texttt{dd}, \texttt{enzymes},
\texttt{nci1}, \texttt{proteins}), 2 time series (\texttt{earth\-quakes}, \texttt{aibo}), 3 text (\texttt{amazon}, \texttt{imdb}, \texttt{yelp}), and 3 image (\texttt{cifar}, \texttt{fashion\-mnist}, \texttt{svhn}) datasets. For the text and image datasets we used embeddings of pretrained RoBERTa\footnote{Liu, Y., Ott, M., Goyal, N., Du, J., Joshi, M., Chen, D., Levy, O., Lewis, M.,
Zettlemoyer, L., Stoyanov, V.: Roberta: A robustly optimized BERT pretraining
approach. \textbf{CoRR abs/1907.11692}, 1–13 (2019)} and ViT\footnote{Dosovitskiy, A., et al.: An image is worth 16x16 words: Transformers for image
recognition at scale. \textbf{CoRR abs/2010.11929}, 1–22 (2020)} models, respectively. 

\textbf{Mixed data.} For mixed-type data, we used 3 sequences of sets (\texttt{item}, \texttt{length}, \texttt{order}) and 3 multiomics (\texttt{ovarian}, \texttt{her2}, \texttt{rosmap}) datasets. Sequences of sets, by nature, can be treated as sets, as sequences, or as combinations of the two representations. The multiomics datasets consist of the results of different genetic measurements, represented as distributions (lengths of variants) or numbers (gene expression).
\clearpage

\begin{figure}[H]
    \centering
    \includegraphics[width=0.75\linewidth]{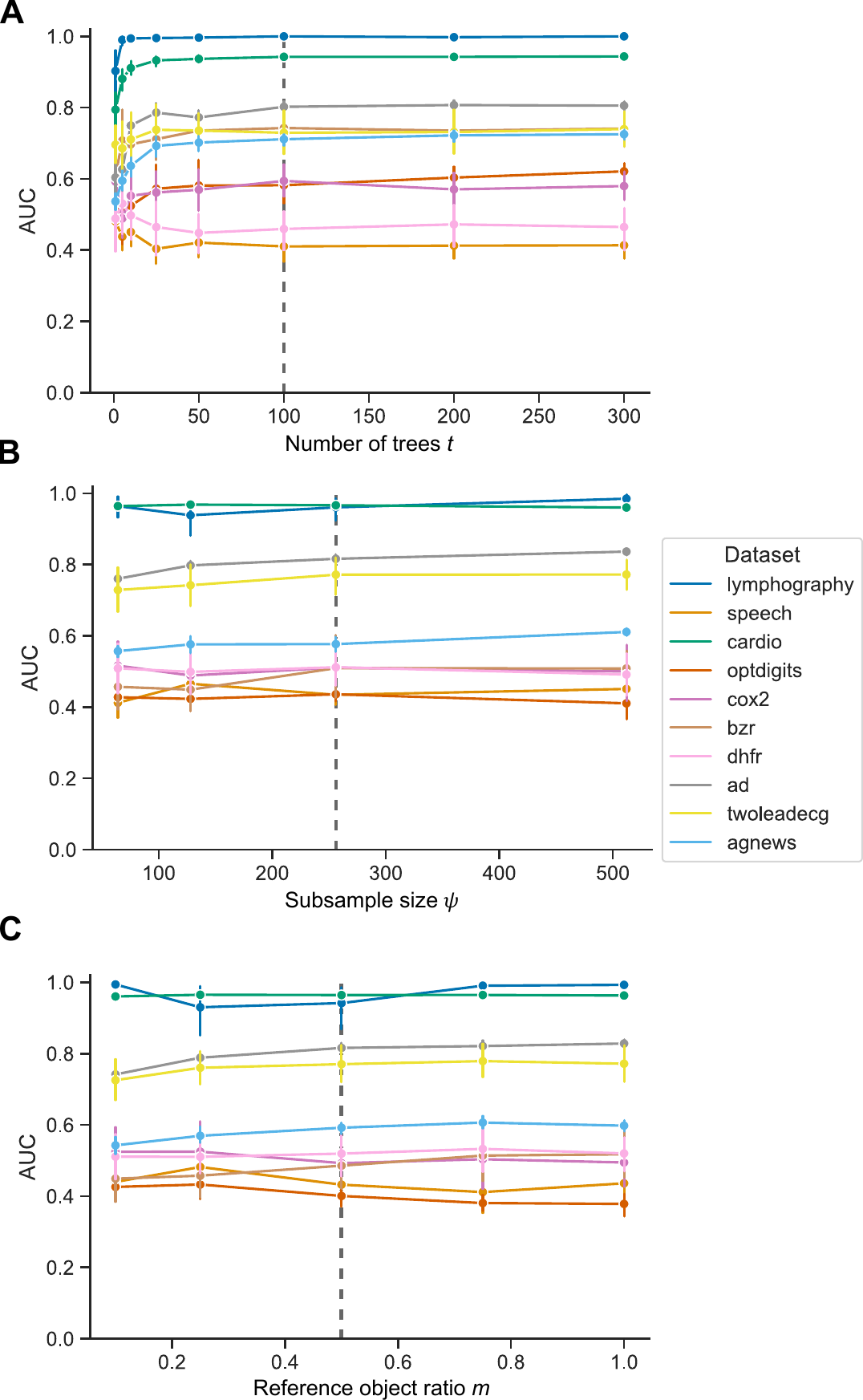}
    \caption{Hyperparameter sensitivity analysis. Effect of (\textbf{\textsf{A}}) the number of trees $t$, (\textbf{\textsf{B}}) subsample size $\psi$, and (\textbf{\textsf{C}}) pool of reference objects on RSIF's predictive performance (ROC AUC). Bars represent 95\% confidence intervals. The dashed gray lines show the selected defaults.}
    \label{fig:hyperparameters}
\end{figure}

\begin{figure}[H]
    \centering
    \includegraphics[width=\linewidth]{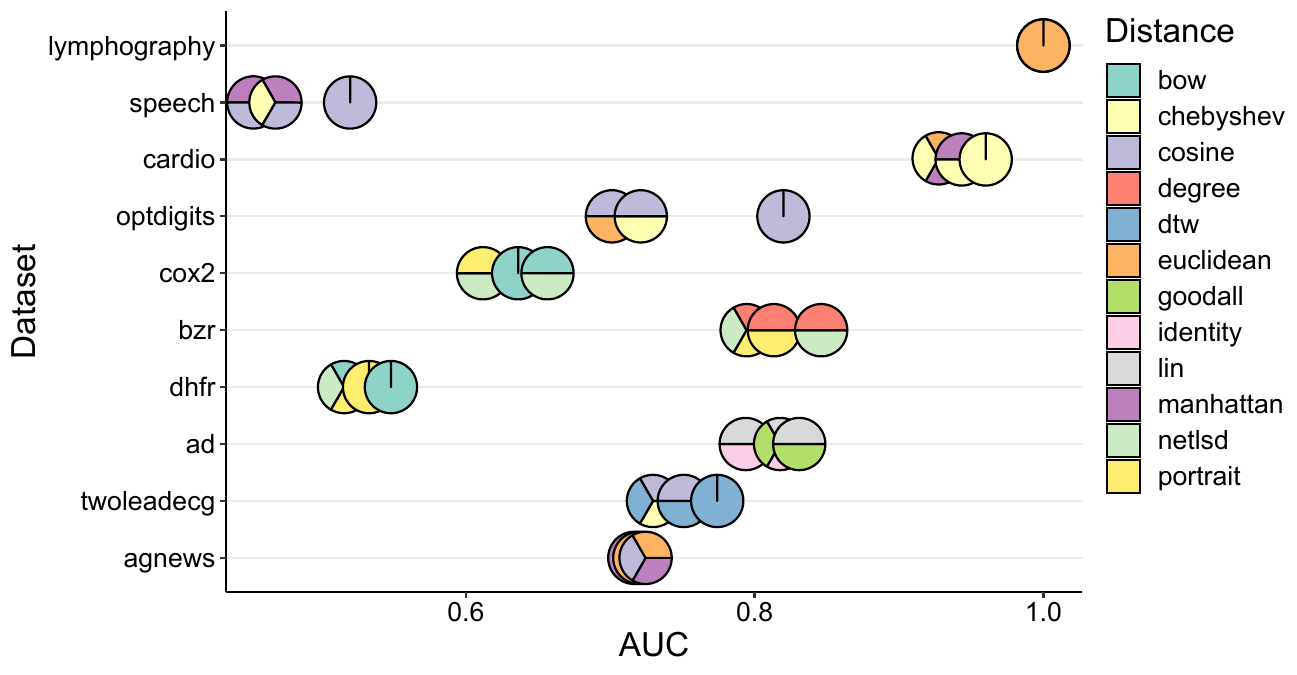}
    \caption{Top 3 distance measure combinations for each of the sensitivity test datasets. Each set of measures used by RSIF is represented by a circle. If the set of distances consists of more than one measure, the circle is divided into multiple colored pieces, with colors defining the measures.}
    \label{fig:distances}
\end{figure}

\begin{figure}[H]
    \centering
    \includegraphics[width=0.65\linewidth]{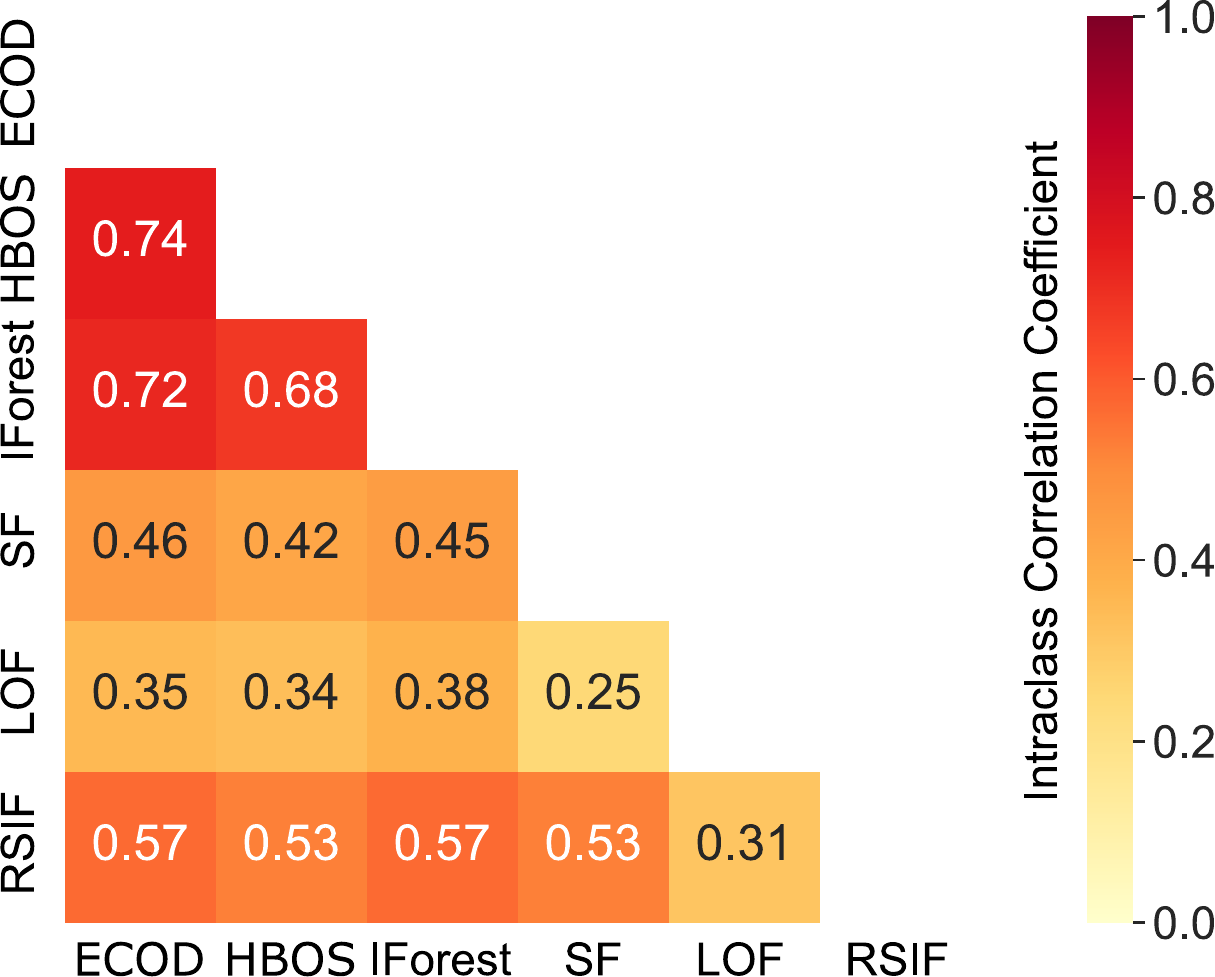}
    \caption{Pairwise average intraclass correlation coefficients (ICC) based on outlier scores for the holdout test sets.}
    \label{fig:icc}
\end{figure}

\begin{table}[htbp]
\scriptsize
\centering
\caption{AUC performance of RSIF and five competitive methods. The best results on each dataset are highlighted in bold, and the second best are underlined.}
\begin{tabularx}{\textwidth}{l@{}ccrrcXXXXXX}
\toprule
\multirow{2.5}{*}{Dataset} & \multirow{2.5}{*}{Type} & \multirow{2.5}{*}{Features} & \multirow{2.5}{*}{\#Ex.} & \multirow{2.5}{*}{\#Feat.} & \multirow{2.5}{*}{\%Outlier} & \multicolumn{6}{c}{AUC}\\
\cmidrule{7-12}
& & & & & & IF & LOF & HBOS & ECOD & {\,\, SF} & RSIF \\
\midrule
\texttt{glass} & \multirow{16}{*}{scalar} & \multirow{11}{*}{numeric} & 214 & 7 & 4.21  &0.72&0.74&\underline{0.76}&0.60&0.73&\textbf{0.80}\\
\texttt{letter} & & & 1600 & 32 & 6.25 & 0.61&\textbf{0.92}&0.59&0.55&0.75&\underline{0.77}\\
\texttt{musk} & & & 3062 & 166 & 3.17 &  1.00&0.60&1.00&0.96&\textbf{1.00}&\underline{1.00} \\
\texttt{annthyroid} & & & 7200 & 6 & 7.42 & \underline{0.80}&0.71&0.61	&0.78&0.68&\textbf{0.84}\\
\texttt{satimage} & & & 5803 & 36 & 1.22 &\underline{0.99}&0.34&0.97&0.96&	0.98&\textbf{0.99} \\
\texttt{thyroid} & & & 3772 & 6 & 2.47 & \underline{0.98}&0.48&0.95&0.98&0.98&\textbf{0.98}\\
\texttt{vowels} & & & 1456 & 12 & 3.43 &  0.69&\textbf{0.93}&0.66&0.59&0.59&\underline{0.91} \\
\texttt{waveform} & & & 3443 & 21 & 2.90 & 0.73&0.74&0.69&0.59&	\textbf{0.83}&\underline{0.76}\\
\texttt{wbc} & & & 223 & 9 & 4.48 &  \textbf{1.00}&0.92&0.99&1.00&0.99&\textbf{1.00} \\
\texttt{wdbc} & & & 367 & 30 & 2.72 & 0.99&0.98&0.99&0.97&\textbf{1.00}&\underline{0.99} \\
\texttt{wilt} & & & 4819 & 5 & 5.33 & 0.46&\textbf{0.69}&0.41&0.39&0.34&\underline{0.53}\\
\cmidrule{3-12}
\texttt{aid} & & \multirow{5}{*}{categorical} & 4278 & 114 & 1.40 &  0.65&0.58&\underline{0.66}&\textbf{0.66}&0.61&0.64 \\
\texttt{apascal} & & & 12694 & 64& 1.39&  0.49&0.55&\underline{0.66}&\textbf{0.66}&0.55&0.56 \\
\texttt{cmc} & & & 1472 & 8 & 1.97&  \underline{0.57}&0.51&\textbf{0.59}&\textbf{0.59}&0.53&\underline{0.57} \\
\texttt{reuters} & & & 12896 & 100 & 1.84&  \underline{0.98}&0.95&\textbf{0.99}&\textbf{0.99}	&\underline{0.98}&\underline{0.98} \\
\texttt{solarflare} & & & 1065 & 11 & 4.04 &  0.80&0.55&\underline{0.84}&0.84&\textbf{0.85}&0.81 \\
\midrule
\texttt{nci1} & \multirow{16.2}{*}{complex} & \multirow{5}{*}{graph} & 2160 & 1 & 4.77&  0.48&\textbf{0.56}&0.46&0.49&0.47&\underline{0.53} \\
\texttt{aids} & & & 1680 & 1 & 4.76 & 0.92&0.83&0.96&0.92&\underline{0.99}&\textbf{0.99} \\
\texttt{enzymes} & & & 105 & 1 & 4.76 &  \textbf{0.76}&0.61&0.68&\underline{0.72}&0.63&0.63 \\
\texttt{proteins} & & & 696 & 1 & 4.47&  0.54&0.58&0.35&0.67&\underline{0.68}&\textbf{0.70} \\
\texttt{dd} & & & 726 & 1 &4.82&  0.66&0.46&0.31&0.75&\textbf{0.79}&\underline{0.78} \\
\cmidrule{3-12}
\texttt{earthquakes} & & \multirow{4}{*}{time series} & 387 & 512 & 4.91 & \underline{0.61}&0.57&0.49&0.56&0.43&\textbf{0.64} \\
\texttt{aibo} & & & 367 & 70 & 4.90 & 0.50&\textbf{0.63}&0.50&0.46&\underline{0.55}	&\underline{0.55} \\
\texttt{ECGFiveDays} & & & 465 & 136 & 4.95 & \underline{0.80}&\textbf{0.91}&0.75&0.67&0.74&0.79\\
\texttt{MPOC} & & & 583 & 80 & 4.97 & \underline{0.68}&\textbf{0.75}&0.62&0.53&0.66&0.61 \\

\cmidrule{3-12}
\texttt{amazon} & & \multirow{3}{*}{text} & 10000 & 768 & 5.00&0.52& \textbf{0.55} & 0.51 &\underline{0.52}&0.49&0.50 \\
\texttt{imdb} & & & 10000 & 768 & 5.00& 0.47&\textbf{0.52}&0.47&0.47&0.48&\underline{0.50} \\
\texttt{yelp} & & & 10000 & 768 &5.00 & 0.54&\textbf{0.59}&0.55&\underline{0.56}&0.50&	0.54 \\
\cmidrule{3-12}
\texttt{cifar} & & \multirow{3}{*}{image} & 5263 & 512 & 5.00& \underline{0.73}&\textbf{0.73}&0.68&0.71&0.68&0.71 \\
\texttt{fashionmnist} & & & 6315 & 512 &5.00& \textbf{0.84}&0.74&0.76&0.83&0.81&\underline{0.83} \\
\texttt{svhn} & & & 5208 & 512 & 5.00&  0.56&\textbf{0.66}&0.48&0.54&\underline{0.57}&0.55 \\
\midrule
\texttt{items} & \multirow{6.4}{*}{mixed} & \multirow{2}{*}{sequences} & 210 & 1 & 4.76 & \underline{0.83}&\underline{0.83}&\textbf{0.84}&\textbf{0.84}&0.75&0.76 \\
\texttt{length} & & \multirow{2}{*}{of sets} & 210 & 1 & 4.76  &  0.85&\underline{0.87}&\textbf{0.92}&\textbf{0.92}&\underline{0.87}&0.81 \\
\texttt{order} & & & 210 & 1 & 4.76 & 0.53&0.53&\underline{0.55}&\textbf{0.59}&0.51&0.54 \\
\cmidrule{3-12}
\texttt{ovarian} & & \multirow{3}{*}{multiomics} & 125 & 50 & 4.80  & 0.50&0.29&0.45&\underline{0.57}&0.33&\textbf{0.69} \\
\texttt{breast} & & & 770 & 50 & 3.64 & 0.62&0.83&0.49&0.63&\underline{0.83}&\textbf{0.83} \\
\texttt{rosmap} & & & 177 & 600 & 4.52 & 0.62&0.60&\underline{0.68}&0.67&\textbf{0.70}&0.66\\
\midrule
&&&&& \textbf{Avg. rank} & 3.50 & 3.57 & 3.78 & 3.51 & 3.78 & \textbf{2.85} \\
\bottomrule
\end{tabularx}
\label{tab:datasets_AUC}
\end{table}

\end{document}